\documentclass[runningheads]{llncs}
\usepackage[T1]{fontenc}
\usepackage{graphicx}
\usepackage{multirow}
\usepackage{caption}
\usepackage{subcaption}
\usepackage{wrapfig}
\usepackage{hyperref}
\usepackage[table,xcdraw]{xcolor}
\usepackage{booktabs}
\usepackage{tabularray}
\usepackage[sectionbib,numbers]{natbib}
\usepackage{floatrow}
\usepackage{amsmath}
\usepackage{rotating}
\usepackage{xcolor}
\usepackage{wrapfig}

\newfloatcommand{capbtabbox}{table}[][\FBwidth]

\def\bpiproblems {\emph{BPI13}}
\def\sepsis {\emph{SEPSIS}}
\def\bpiseventeen {\emph{BPI17}}
\def\bpipermit {\emph{BPI20}}
\def\ltlf {$\text{LTL}_f$}

\begin{document}
\title{CoSMo: a Framework to Instantiate Conditioned Process Simulation Models} 
\titlerunning{CoSMo}
%
\author{Rafael S. Oyamada\inst{1} \orcidID{0000-0002-4408-9575} 
\and
Gabriel M. Tavares\inst{2, 3} \orcidID{0000-0002-2601-8108} 
\and
Sylvio Barbon Junior\inst{4} \orcidID{0000-0002-4988-0702}
\and
Paolo Ceravolo\inst{1} \orcidID{0000-0002-4519-0173}
}
\authorrunning{Oyamada et al.}

\institute{
Universit\`a degli Studi di Milano, Milan, Italy\\
\email{\{rafael.oyamada,paolo.ceravolo\}@unimi.it} \and
LMU Munich, Munich, Germany\\
\email{tavares@dbs.ifi.lmu.de}\and
Munich Center for Machine Learning (MCML), Munich, Germany \and
Universit\`a degli Studi di Trieste, Trieste, Italy\\
\email{sylvio.barbonjunior@units.it}\\
}
\maketitle
\begin{abstract} 
Process simulation is gaining attention for its ability to assess potential performance improvements and risks associated with business process changes. 
The existing literature presents various techniques, generally grounded in process models discovered from event log data or built upon deep learning algorithms.
These techniques have specific strengths and limitations. 
Traditional data-driven approaches offer increased interpretability, while deep learning-based excel at generalizing changes across large event logs. 
However, the practical application of deep learning faces challenges related to managing stochasticity and integrating information for what-if analysis.
This paper introduces a novel recurrent neural architecture tailored to discover COnditioned process Simulation MOdels (CoSMo) based on user-based constraints or any other nature of a-priori knowledge.
This architecture facilitates the simulation of event logs that adhere to specific constraints by incorporating declarative-based rules into the learning phase as an attempt to fill the gap of incorporating information into deep learning models to perform what-if analysis.
Experimental validation illustrates CoSMo's efficacy in simulating event logs while adhering to predefined declarative conditions, emphasizing both control-flow and data-flow perspectives.

\end{abstract}

\section{Introduction}

Process simulation models are instrumental in a variety of Process Mining (PM) applications including compliance analysis, performance analysis, and what-if analysis. 
Predominantly, these models are categorized into Data-driven Process Simulation (DDS) models~\cite{MartinDC16,CamargoDR19b,CamargoDG20,BurattinRRT22}, Deep Learning-based (DL) solutions~\cite{CamargoDR19,CamargoDR21}, and hybrid approaches combining both~\cite{CamargoDR22hybridsim,MeneghelloFG23rims}, each offering unique capabilities and facing distinct challenges. 
While DDS models excel in simulating control-flow patterns, they are often hindered by oversimplified assumptions about resource behavior and are susceptible to biases from event log characteristics. 
Conversely, DL models, despite their success in predictive process monitoring~\cite{ManeiroVL21}, struggle with the stochastic nature and the challenges of integrating what-if scenario knowledge, limiting their utility in process simulation~\cite{Dumas21}.
A natural aspect common to both DDS and DL models is the stochastic approach in simulation, relying on randomness to yield realistic outcomes~\cite{Aalst15}.
While this enhances realism, it might limit user control over simulated behavior and hinder practical flexibility (e.g., for what-if analysis).
In this sense, DDS models overcome this issue by providing tunable parameters~\cite{CamargoDR19b}.
On the other hand, DL-based process simulation models typically do not provide this flexibility.
These models are often seen as black boxes, with no options for users to influence the simulation results, relying solely on random sampling methods to draw predictions from the probability distributions outputted by the model.

Therefore, in this paper, we motivate the use of deep learning for process simulation due to its successful accomplishments in other process mining tasks, and we address the problem associated with the lack of flexibility of current DL-based solutions for process simulations~\cite{Dumas21}. By proposing a novel methodology to introduce user-directed controls and guidance within these models, our work advances the fields by introducing a new bridge between advanced process simulation capabilities using deep learning and practical usability needs.
Thus, we comprise our contribution as follows: 
(i) a novel conditioned recurrent architecture tailored to capture the relationship between user-defined constraints and the sequential nature of events; 
(ii) the introduction of a new framework, \textbf{Co}nditioned Process \textbf{S}imulation \textbf{Mo}del (CoSMo), which implements the conditioned recurrent architecture using any type of constraint or a-priori knowledge; 
(iii) the validation of this approach through experiments considering both control-flow and data-flow perspectives. 
CoSMo has proven its capability to simulate traces in compliance with constraints represented by the DECLARE language model~\cite{PesicSA07declare}, providing users with significant control over their generative process models and effectively overcoming the stochastic limitations inherent in traditional DL-based process simulation models. 

This paper details the development of CoSMo and its evaluation, highlighting its alignment with current literature and its advancements to traditional process simulation approaches.
The remainder of this paper is structured as follows: Section~\ref{sec:background} provides background information. 
Section~\ref{sec:related} discusses the current state of process simulation and its limitations.
Section~\ref{sec:proposal} introduces CoSMo, our proposed framework for implementing process simulation models based on user-based constraints or any other nature of a-priori knowledge available.
Section~\ref{sec:evaluation} describes our evaluation and limitations.
Finally, Section~\ref{sec:conclusion} concludes the paper and outlines future work.

\section{Background}\label{sec:background}

An event log $L$ consists of a set of cases $Q$ (a.k.a. process executions). 
Each case $q_i \in Q$ is composed of, essentially, a trace of events $t_i$ and a case identifier $i$. 
A trace is composed of a finite ordered sequence of $s$ events and each event $e_{i,j}$ refers to the j-th execution, for $0 \leq j \leq s$, of a system activity in the trace $t_i$ and is characterized by a set of alternative attributes.
Thus, an arbitrary event $e$ can be represented as a tuple $e = (i, a, time, (d_1, v_1), \dots, (d_z, v_z))$, where $a \in A$ is the activity label from the set of activities present in $L$, $time$ is the timestamp denoting when the activity was executed, and $(d_1, v_1), \dots, (d_z, v_z)$ are the event or case attributes and their values, with $z \geq 0$.

A process model $N$ is an abstract representation capturing the sequence and interrelation of activities in historical process executions, as recorded in an event log. 
Process models can be classified as either imperative, like BPMN~\cite{RosingWCM15bpmn}, which define specific relationships of activities, or declarative, like the DECLARE model~\cite{PesicSA07declare}, which set rules and constraints without specifying exact sequences.
On the other hand, process simulation models are defined as a set of techniques that aim at stochastically predicting possible behaviors of a process~\cite{Aalst15}. 
It might be usually defined as a tuple $M=(N, P)$, where $N$ is a discovered process model (e.g., a BPMN model) and $P$ is a set of parameters needed to define the simulation specifications for the different process perspectives, such as the control-flow, the resource, and time perspectives. 
These parameters can be manually defined based on expert knowledge~\cite{RozinatMSA09} or automatically extracted from the event log~\cite{CamargoDR19b}. 

We now provide essential definitions related to the DECLARE model used in this work.
The process modeling language DECLARE~\cite{PesicSA07declare} is based on Linear Temporal Logic over finite traces (\ltlf)~\cite{GiacomoV13ltl}, which is used to express process semantics. 
A DECLARE model is defined by a set of activities and constraints on these activities. 
These constraints are formalized using \ltlf~formulae. 
DECLARE utilizes a subset \ltlf~formulae, embodied in predefined patterns known as DECLARE templates. 
Following the notations presented by Donadello et al.~\cite{Donadello23oodeclare}, we define the complete set of DECLARE templates as $\mathcal{DT} = \mathcal{E} \cup \mathcal{C} \cup \mathcal{PR} \cup \mathcal{NR}$, where each template is defined as follows.
\textbf{Existence} ($\mathcal{E}$): given an arbitrary activity $a$, this group of templates checks the number of occurrences, its absence, or its position in the trace;
\textbf{Choice} ($\mathcal{C}$): given $a$ and $b$, this group of templates check if $a$ or $b$ occurs, or if $a$ and $b$ occur together;
\textbf{Positive Relations}($\mathcal{PR}$): given $a$ and $b$, this group checks the relative position of the activities. For instance, if $a$ occurs, $b$ eventually occurs;
\textbf{Negative Relations}($\mathcal{NR}$): given $a$ and $b$, this group checks if the activities do not occur together or do not occur in a certain order. For instance, if $a$ occurs, $b$ never occurs.
In the context of this work, we omit further details and formal definitions as we follow the same as defined in the literature~\cite{Donadello23oodeclare,AlmanMMP22pdeclare}. 


\section{Current State and Limitations of Process Simulation}\label{sec:related}

Process simulation models aim at abstracting details from the event logs in order to simulate reality~\cite{Aalst15}.
They are employed as a tool by the PM community for several applications, such as conformance checking~\cite{SaniGZA20}, event log generation~\cite{Burattin15,CamargoDR21}, purposed-oriented event log generation~\cite{BurattinRRT22}, what-if analysis~\cite{Dumas21,CamargoDR22hybridsim}, and process monitoring~\cite{MeneghelloFG23rims}.

DDS models are built upon process models discovered from event logs~\cite{RozinatMSA09,Burattin15,BurattinRRT22}, and popular proposals include PGL2~\cite{Burattin15} and SIMOD~\cite{CamargoDR19b}. 
These methods are typically designed to simulate different control-flow patterns by manipulating user-based requirements such as the number of gates and the amount of noise.
However, DDS models are constrained by oversimplified assumptions about resource behavior, neglecting complexities such as multitasking, batching, and resource sharing~\cite{Estrada21ddslimitation}. 
Despite the outstanding performance of deep learning models in predictive monitoring~\cite{TaxVRD17,Francescomarino17,Pasquadibisceglie19,CamargoDR19,ManeiroVL21}, its use in process simulation is limited by its stochastic nature and difficulty in adapting to what-if scenarios~\cite{CamargoDR21}, often leading to the limited of DL as generative models~\cite{CamargoDR21} or hybrid models with process mining techniques~\cite{CamargoDR22hybridsim,MeneghelloFG23rims}.

A common characteristic among process simulation models is the stochasticity inserted during simulation.
The simulation process requires a sampling step to draw information from the probability distributions, and applying a certain level of randomness helps models produce more realistic simulations~\cite{Aalst15}.
For example, DDS-based simulations are supported by statistical measures extracted from the event logs, such as branch probabilities and activity duration time distributions. 
On the other hand, DL-based approaches consist of learning the underlying data distributions and drawing event attributes from the probability distributions returned by the learned model. 
Although this stochastic approach makes simulations more realistic~\cite{Aalst15,CamargoDR19}, the complete dependence on randomness limits the flexibility of the user to control the simulated behaviors~\cite{Dumas21}.

To illustrate this stochastic problem, let us consider the problem of next activity prediction in PM~\cite{ManeiroVL21}. 
The goal here is to always determine the most likely next activity in an ongoing process. 
Conversely, in process generation or simulation, diversity is key because realistic process traces typically vary. 
Relying solely on the most likely sequence results in monotonous, identical trace variants. 
Typically, this problem is addressed by sampling the next activity from a probability distribution, using techniques such as multinomial sampling~\cite{TaxVRD17} or beam search~\cite{Francescomarino17}. 
However, such stochastic methods reduce the user's ability to precisely control the simulation, especially when performing what-if analyses.

To address the inherent stochasticity in deep learning models, we introduce an innovative recurrent architecture specifically designed to recognize the relationship between the sequential nature of events and various forms of a-priori knowledge, articulated as user-provided constraints. 
In this paper, we use the DECLARE model to extract semantics from event logs and reuse them as constraints for learning and compliance. 
Our methodology is informed by previous research that has effectively used the DECLARE language to enrich or encode training data~\cite{Francescomarino17,Donadello23oodeclare}. 
By extending these concepts to the simulation domain, our approach provides users with greater control over their generative process models and effectively mitigates stochastic challenges. 
This strategy not only aligns with the current literature but also pushes the boundaries of conventional modeling techniques, promising significant advances in the field. 
In the following sections, we further describe our proposal as well as the experimental evaluation of our approach.

\section{Conditioned Process Simulation Models}\label{sec:proposal}

\begin{figure}
  \centering
  \begin{subfigure}[b]{0.57\textwidth}
      \centering
      \includegraphics[width=\textwidth]{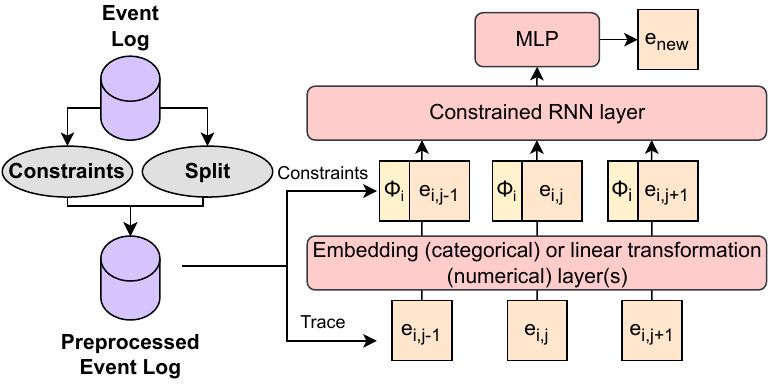}
      \caption{CoSMo framework overview.}
      \label{fig:cosmo}
  \end{subfigure}
  \hfill
  \begin{subfigure}[b]{0.39\textwidth}
      \centering
      \includegraphics[width=\textwidth]{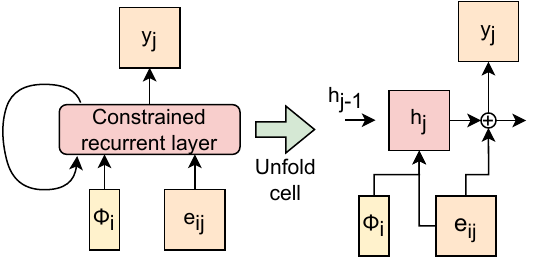}
      \caption{Proposed conditioned recurrent layer.}
      \label{fig:cosmo-rnn}
  \end{subfigure}
  \caption{Proposed CoSMo framework (left) and the proposed recurrent architecture (right) tailored to capture the relation between user-based constraints and processes.}
  \label{fig:cosmo-overview}
\end{figure}

In this section, we introduce CoSMo, our proposed framework for implementing process simulation models based on user-based constraints or any other type of a-priori knowledge available. 
In a nutshell, given an event log, the instantiation pipeline of our framework consists of a preprocessing phase, where the user will split and extract their custom constraints, and train our proposed conditioned recurrent network. 
This pipeline is depicted in Figure~\ref{fig:cosmo} and it will be better described in this section. 
We first introduce the recurrent network tailored to learn how to generate data based on constraints, then we explain how this network can accept any constraint, and finally, we illustrate the example of conditioning the simulation of processes based on \ltlf~rules.


\subsection{Conditioned Recurrent Architecture}

Our proposed architecture starts by taking a sequence of events $x$ as input, where each event element is composed of different attributes, as described in Section~\ref{sec:background}. 
Alongside this sequence, the architecture is introduced with external feature vectors $c$ representing a set of constraints. 
We extend the vanilla recurrent neural network (RNN) by introducing a new learnable parameter $K$ to capture the relation between the condition $\Phi$ and the sequential input, followed by a residual connection between each recurrent cell:


\begin{equation}
    \begin{gathered}
        h_j=f(Ux_j+Wh_{j-1}+K\Phi) \\
        y_j=g(Vh_j) + x_j
    \end{gathered}
\end{equation}


Where $h_j$ is the hidden state at time step $j$. This represents the ``memory'' of the network up to that point in time.
On the other hand, $U$, $W$, and $V$ are the learnable parameters from the vanilla RNN, and each plays a distinct role in transforming the input and hidden states. 
The parameter $K$ specifically learns how the constraint feature vector $\Phi$ relates to the input sequence at every timestep. 
The key for the proposed recurrent network is that these condition vectors are external to the primary sequence but highly relevant to the task at hand. 
In simple terms, a basic RNN learns a hidden state for each event in the sequence. 
Our model introduces extra parameters, $K$, which act like an additional hidden state. 
This extra state learns the relationship between the set of rules at each time step and its corresponding hidden state. 
Finally, $g$ is a non-linear activation function, essential for enabling the network to capture complex patterns and the residual connection at $y_j$ helps to improve and stabilize the learning phase. 
This residual technique is common in other deep learning applications and modern recurrent architectures, such as LSTMs, have a similar characteristic (e.g., the forgetting gate mechanisms). 
The bias terms are omitted for simplicity. 
We depict the new architecture in Figure~\ref{fig:cosmo-rnn}.


Ultimately, given an input sequence $x$ and a constraint vector $\Phi$, our proposed backbone layer learns the conditioned probability $p(y | x, \Phi) = \hat{y}$ to predict the next attribute of that sequence. 
Theoretically, these constraint features can take any form, such as generic user-based constraints, intra-case features describing the availability of shared resources at that timestamp, or declarative rules based on \ltlf. 
In this paper, we dive into the latter and demonstrate how the simulation can be guided by enforcing the generation of traces that satisfy different DECLARE templates of \ltlf~rules.

\subsection{CoSMo Instantiation}

In this section, we describe how to instantiate CoSMo by taking into consideration the \ltlf~rules extracted from the event log using the set of DECLARE templates $\mathcal{DT}$ described in Section~\ref{sec:background} and the conditioned recurrent architecture previously described. 
We highlight that any other nature of constraints can be employed in our framework, but we focus on \ltlf~rules in this paper.

Consider an event log $L$ of $n$ traces and a group of DECLARE templates, where this group can be represented by $dt$ from the set of DECLARE templates of \ltlf~rules described in Section~\ref{sec:background}, i.e., $dt \in \mathcal{DT}$.
For every trace $t_i$ in the event log $L$, the DECLARE model discovers a vector $\Phi_i = \lbrack \phi_0, \dots, \phi_m \rbrack $ of $m$ \ltlf~rules from $dt$ that are fulfilled for $t_i$.
The original log $L$ can be written as $L=\lbrace{t_i}\rbrace_{i=0}^n$.
This log is then augmented to form a new log $L'$, which is defined as $L'= \{\langle t_{i}, \Phi_i \rangle \}_{i=0}^n$.
Here, $\Phi$ is represented as a binary feature vector of $m$ dimensions (constraints) that describes if the $\hat{m}-th$ constraint is fulfilled in $t_i$ or not, for $0 \leq \hat{m} \leq m$.

Subsequently, the event log $L'$ is split into train and test sets ($L'_{train}$ and $L'_{test}$, respectively).
The training phase is performed by feeding the network with a trace $t_i$ containing $s$ events $t_i=[e_0, ..., e_s]$ and its respective constraint vector $\Phi_i$. 
As depicted in Figure~\ref{fig:cosmo}, the $\Phi_i$ vector is concatenated to each event vector $e_{i,j} \in t_{j=0}^s$ in order to enforce the model to learn the relation between the set of rules and the sequential nature of events. 
The network is trained to predict the next event $e_{new}$ and the loss function is computed by comparing the predicted event attributes with the ground truths. 
This process constitutes a multitasking framework, especially when there are multiple event attributes to be predicted. 
In this paper, we focus on the process simulation by predicting in an autoregressive fashion the next activity label and the remaining time of the trace based on a set of constraints.

\subsection{Conditioned Simulation}

After instantiating and training the CoSMo model, we initiate the conditioned simulation of new traces.
This process begins by selecting constraint vectors, $\Phi_{test}$ from the test dataset $L'_{test}$.
For clarity, we will drop the `test' subscript in this explanation.
The core idea of our approach is to modify specific \ltlf~rules, $\phi$, within the set $\Phi$, thereby directing CoSMo to create new traces based on these modifications.
The goal is to control the process behavior by specifying which rules are to be fulfilled or not in the generation of a new trace.
Thus, from the set $\Phi$ we select a subset of rules $\Phi' \subset \Phi$, and invert their values to condition the simulation of process behaviors.
This selective modification, as opposed to altering all rules, is crucial to maintaining a diversity of trace behaviors;
otherwise, using the same full set $\Phi$ would result in repetitive, identical behaviors.
For instance, to enforce the existence of an activity $a$, an arbitrary subset $\Phi'$, which is defined as a binary vector, e.g. $\Phi' = [existence(a)=0]$, has its value inverted.
Therefore, after inverting the values from $\Phi'$ and given the preserved subset of rules $\Phi^p = \Phi - \Phi'$, we design a new set of constraints $\Phi^s = \Phi^p \cup \Phi'$ to perform our conditioned simulations. 
Starting with an arbitrary initial event $e_0$, we autoregressively generate subsequent events, each conditioned on the set $\Phi^s$, thereby simulating the process according to these newly defined constraints.

\textbf{Example.}
Consider an event log composed of traces $L = \{<a,b,c>, <a,a,c>\}$ and the instantiated (trained) CoSMo framework.
Considering the usage of the Existence template, assume the DECLARE model identifies the following constraints: $\Phi = \{ \text{Existence}(a), \text{Existence}(a), \text{Existence}(c) \}$.
In practice, $\Phi$ is represented as a matrix $n \times m$, where $n$ is the number of traces and $m$ is the number of found constraints. 
For example:
$\Phi = \begin{bmatrix} 1 & 1 & 1 \\ 1 & 0 & 1 \end{bmatrix}$.
Here, $1$ indicates that the constraint is met, and $0$ indicates that it is not.
Next, let us define the subset of constraints $\Phi' = \{ \text{Existence}(c) \}$ to be inverted. 
Thus, the final set of constraints $\Phi^s$ to be simulated will be:
$\Phi^s = \begin{bmatrix} 1 & 0 & 1 \\ 1 & 1 & 1 \end{bmatrix}$, where the second column values are inverted.
Given an arbitrary first event $e_0 = [a]$, let us assume the model needs to generate the next event $e_1$ based on the arbitrary constraints $\phi = [1, 1, 0]$ (in simple terms, $a$ and $b$ should exist, whereas $c$ should not).
The generated $e_1$ will be concatenated to the input sequence and this new sequence will be used to generate the next event $e_2$.
Each of these events will always consist of an activity label and its execution time.
The model continues this autoregressive process, adding each generated event to the input trace and predicting the next event based on the imposed constraints.
The stop criterion consists of defining the maximum trace length or when a specific event label is generated.
Examples of valid traces that satisfy the set of constraints include $<a,b>$ and $<a,a,b>$, among others.

\section{Evaluation}\label{sec:evaluation}

In the following section, we delineate our experimental framework, illustrate its application through a use case involving real-world event logs for the assessment of our proposed methodology, and conclude with an analytical discourse on the prevailing limitations and strengths of our approach.

\subsection{Experimental Setup}\label{sec:setup}

\textbf{Event logs}. We employ five distinct real-world datasets, each with unique characteristics, sourced from the 4TU.ResearchData repository. To ensure a thorough yet focused evaluation, we attempt to choose datasets of varying characteristics (number of unique activities, number of events, a trace length) -- small, medium, and large, illustrated in Table~\ref{tab:log-properties}. 
Some of these event logs have different versions. In this work, more precisely, 
we use the complete \emph{BPI12}\footnote{\url{https://data.4tu.nl/articles/_/12689204/1}}, 
the problems version of \bpiproblems\footnote{\url{https://data.4tu.nl/articles/_/12688556/1}},
and the travel permit subset of \bpipermit\footnote{\url{https://data.4tu.nl/articles/_/12718178/1}}.
The \bpiseventeen\footnote{\url{https://data.4tu.nl/articles/_/12696884/1}}  
and\sepsis\footnote{\url{https://data.4tu.nl/articles/_/12707639/1}} 
datasets are used in their entirety.

\begin{table}[]
  \centering
  \scriptsize
  \resizebox{.6\columnwidth}{!}{%
  \begin{tabular}{@{}c|c|c|c|c@{}}
    \toprule
    \textbf{Event Log} & \textbf{Acts.} & \textbf{Evts. ($10^3$)} & \textbf{Cases ($10^3$)} & \textbf{C. Length} \\ \midrule
    BPI12  & 23 & 104.82  & 9.49  & 11.05 ± 9.64  \\
    BPI13  & 6  & 4.89    & 1.33  & 3.69 ± 4.09   \\
    BPI17  & 26 & 1210.81 & 31.50 & 38.44 ± 17.96 \\
    BPI20  & 50 & 82.22   & 6.85  & 12.01 ± 5.46  \\
    SEPSIS & 8  & 9.87    & 0.95  & 10.37 ± 3.9   \\ \bottomrule
  \end{tabular}%
  }
  \caption{Event logs and their properties: number of unique activities, number of events, number of cases, and average case length.}
  \label{tab:log-properties}
  \end{table}

\textbf{Preprocessing}. Our preprocessing phase is structured into the following stages: initial data cleaning, extraction of a-priori knowledge, and then dividing the data into training and testing sets.
This initial stage involves removing traces that are either shorter than three events or longer than the 90th percentile in the trace lengths' distribution, a strategy that ensures a more even distribution of sequence lengths. 
Following the cleaning process, we use the Declare4Py library~\cite{Alman24declare4py} to extract the a-priori knowledge by discovering a DECLARE model from the event log, utilizing the discovered \ltlf~rules as constraints to feed our proposed deep neural net. 
For the datasets \emph{BPI12}, \bpiseventeen, and \bpipermit, we employ the benchmarked splits~\cite{WeytjensW21a}, while for the \bpiproblems~and \sepsis~datasets, the data is split using a publicly available implementation from the popular PM4PY library\footnote{\url{https://pm4py.fit.fraunhofer.de/}}.

\textbf{Architecture design and training}. This paper introduces a conditioned recurrent network model that integrates categorical and numerical event attributes, enriched with preprocessing constraints, as its input (see Figure~\ref{fig:cosmo-overview}). 
A Multi-Layer Perceptron (MLP) is employed upon the proposed conditioned backbone, and it operates at the final stage to generate new events, symbolized as \(e_{new}\). 
The model performs multitasking by autoregressively generating multidimensional events, utilizing previous predictions of activity labels and execution times to forecast subsequent labels and times. 
Specifically, it predicts the next activity label and the remaining time, where execution time is the interval between the current and previous event timestamps, and the remaining time is the expected duration to complete the process. 
Model evaluation employs cross-entropy and mean squared error for loss functions, addressing categorical and numerical predictions respectively. 
Hyperparameter tuning involved grid search for optimal learning rates \([5e-3, 1e-3, 5e-4, 1e-4]\), number of epochs \([50, 100]\), batch sizes \([16, 32, 256, 512]\), and recurrent layer configurations including input sizes \([16, 32, 64, 128, 256]\), hidden sizes \([32, 64, 128, 256, 512, 1024]\), and layer counts \([1, 2, 4]\).
Due to the lack of space, we do not include the complete discussion on the hyperparameter tuning results, but we report the best configuration.
The Python implementation is available in our repository~\footnote{\url{https://github.com/raseidi/cosmo}}.

\textbf{Simulation}. Building upon the methodology outlined in Section~\ref{sec:proposal}, we initiate by selecting the constraints $\Phi_{\text{test}}$, represented in our study as groups of DECLARE templates. 
To condition the simulation, we then identify a subset $\Phi'$ from these constraints to be inverted.
Given the frequency distribution of the set of rules, a pair of activities (or a single activity when the template is the Existence) is drawn from it taking into consideration only the rules lying between the 1st and 3rd quantile of that distribution.
By setting this boundary, we aim at not selecting rules with low variance in order to accomplish greater generalization.
Despite the arbitrary intuition behind this selection, in the sense of business-driven choices, we propose this data-driven approach to ensure a fair evaluation across all event logs and eliminate the need for an expert to define the rules to be simulated for each event log.
Although this is a clear limitation in real-world scenarios, it effectively validates our scientific methodology. 

Subsequently, in order to ensure consistency in our evaluation across all event logs, we uniformly select the same DECLARE templates. 
As previously mentioned, these rules are expressed as vectors of binary values, where `1' indicates rule fulfillment, and `0' denotes non-fulfillment. 
Our primary aim is to invert one specific subset of rules to `enforce' desired process behavior. 
However, since we are using all the available DECLARE templates, we must ensure that the rules are not contradictory.
For instance, inverting the `existence' rule while maintaining the `absence' rule could create contradictory conditions, leading to a conflict in the model's inference process. 
We perform a careful selection of templates to avoid any inconsistencies: we do not include the $\mathcal{NR}$ template group in our evaluation, as it is inversely related to the $\mathcal{PR}$ group, in general.
Such meticulous template selection ensures coherence and efficacy in our model's simulation capabilities.
Thus, if we want to ensure the `existence' rule, the `absence' must be set as the opposite as well. 

Therefore, for each group of templates, we pick the following rules as references: 
the Existence(a) regarding the group $\mathcal{E}$, the Exclusive choice(a,b) regarding the group $\mathcal{C}$, and the Chain Response(a, b) regarding the group $\mathcal{PR}$.
Subsequently, the following rules are also picked to avoid the aforementioned inconsistency issue: (i) for group $\mathcal{E}$, the templates Absence(a) and Exactly(1, a) are chosen; (ii) for group $\mathcal{C}$, no extra template is selected; and (iii) for group $\mathcal{PR}$, we select the Alternate Response(a, b), Precedence(a, b), and Response(a, b) templates.
The remaining rules are preserved ($\Phi^p$) and the picked ones are inverted ($\Phi'$) to compose the final set of constraints $\Phi^s=\Phi^p \cup \Phi'$ to be simulated. 
Each of these templates and its details is described in~\cite{Donadello23oodeclare}.
We selected the activities based on their frequencies (we picked the 3rd quartile), and the activities incorporated into each group of templates for each event log are described in Table~\ref{tab:simulated-templates}.

\begin{table}[]
  \centering
  \resizebox{\columnwidth}{!}{%
  \begin{tabular}{@{}cccc@{}}
  \toprule
  \textbf{Event Log} & \textbf{Template}  & \textbf{Activity A}                  & \textbf{Activity B}                  \\ \midrule
  \multirow{3}{*}{BPI12}  & Existence          & A\_DECLINED                          & -                                    \\
                          & Choice             & W\_Completeren aanvraag              & W\_Nabellen offertes                 \\
                          & Positive Relations & A\_PARTLYSUBMITTED                   & W\_Completeren aanvraag              \\ \midrule
  \multirow{3}{*}{BPI13}  & Existence          & Queued-Awaiting Assignment\_complete & -                                    \\
                          & Choice             & Accepted-Assigned\_complete          & Queued-Awaiting Assignment\_complete \\
                     & Positive Relations & Accepted-In Progress\_complete       & Queued-Awaiting Assignment\_complete \\ \midrule
  \multirow{3}{*}{BPI17}  & Existence          & W\_Handle leads                      & -                                    \\
                          & Choice             & A\_Submitted                         & A\_Cancelled                         \\
                          & Positive Relations & O\_Accepted                          & A\_Pending                           \\ \midrule
  \multirow{3}{*}{BPI20}  & Existence          & Permit APPROVED by BUDGET OWNER      & -                                    \\
                     & Choice             & Declaration APPROVED by BUDGET OWNER & Declaration SUBMITTED by EMPLOYEE    \\
                          & Positive Relations & End trip                             & Send Reminder                        \\ \midrule
  \multirow{3}{*}{SEPSIS} & Existence          & LacticAcid                           & -                                    \\
                          & Choice             & Admission NC                         & LacticAcid                           \\
                          & Positive Relations & Admission NC                         & CRP                                  \\ \bottomrule
  \end{tabular}%
  }
  \caption{DECLARE templates and their respective activities employed to perform the conditioned simulation.}
  \label{tab:simulated-templates}
  \end{table}

\textbf{Evaluation and Validation}. Upon generating a simulated event log under these inverted constraints, we employ conformance checking to verify the fulfillment of the modified rule set.
In this evaluation, we split the traces between two classes to quantify the effectiveness of the simulation: one where the reference rule must be fulfilled and the other where it must not.
For instance, a trace where the reference rule must be fulfilled (e.g., Existence(a)=1) will be a true positive if the simulation conforms to the rule, otherwise, it will be a false negative. 
The performance is then reported through the calculation of weighted precision, a metric chosen specifically for its ability to provide nuanced insight into the degree to which the simulated traces adhere to or deviate from their intended behavior. 
The use of weighted precision is particularly appropriate in this context because it allows for a nuanced assessment of each trace's adherence to the imposed rules, recognizing the varying importance of different rules within the process, thereby providing a more granular and meaningful measure of CoSMo's effectiveness and ability to replicate or alter specific process behaviors as needed.

In addition, we include in this evaluation two baselines against the aforementioned inverted subset. 
First, we compare CoSMo with this perturbed set of constraints ($\Phi^s$) as input to CoSMo with the original set of constraints ($\Phi$) as input, which is the simplest way to assess whether the model actually changes or maintains the input set of rules.
Simulating the \emph{Original} and \emph{Inverted subset} of rules is an alternative to performing an ablation study, which is a common practice in the machine learning literature to assess the impact of a particular component in a model.
In practical terms, by providing the \emph{Original} set of rules, we expect to validate the model's capability to learn the relationship between those rules and the process semantics.
On the other hand, the \emph{Inverted subset} allows us to validate the model's capabilities to generalize.
More specifically, this approach could be thought of as the changes that we aim to do in a process, which could characterize a what-if scenario. 
Second, we compare CoSMo to a \emph{Vanilla} simulation without any constraints, which is the most common way to stochastically generate processes using deep learning in the literature~\cite{CamargoDR21}. 
The intuition behind this comparison is to assess whether the model is actually learning the constraints and whether the constraints are affecting the simulation. 
We use the architecture introduced in~\cite{TaxVRD17} as a vanilla baseline, since it overcomes other approaches from the predictive monitoring literature~\cite{ManeiroVL21}.
We run each simulation 10 times to ensure the robustness of the results since the stochastic nature of the model can lead to different results in each run.

We then discuss the capabilities and limitations of adapting complex \ltlf~constraints by taking into account data and process characteristics.
Finally, we present the remaining time prediction performance through the mean absolute error (MAE) of the test set obtained during training, a common metric in the predictive monitoring literature~\cite{ManeiroVL21}.

\subsection{Conditioned Simulation of Process Behaviors}

\begin{figure}
  \centering
  \includegraphics[width=.8\textwidth]{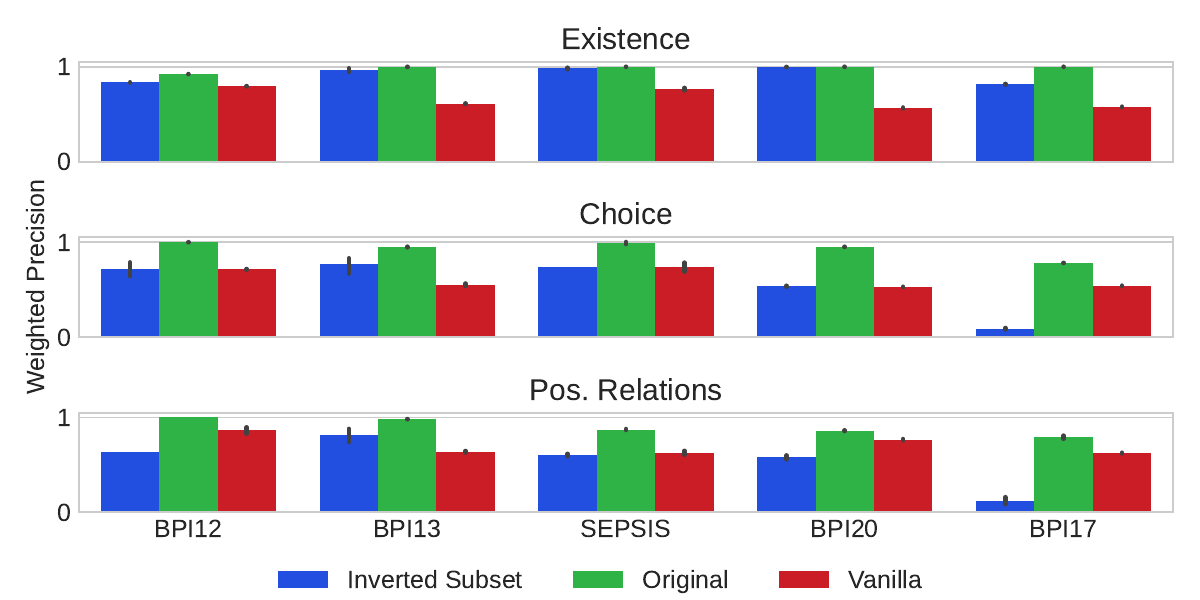}
  \caption{Weighted precision for the simulated logs under the different approaches and DECLARE templates.}
  \label{fig:weighted-precision}
\end{figure}

\textbf{Precision evaluation}. Figure~\ref{fig:weighted-precision} illustrates the weighted precision of the simulated logs under the different approaches.
The \emph{Vanilla} is the stochastic model without any constraints, the \emph{Original} is the model with the original set of constraints, and the \emph{Inverted subset} is the model with the inverted set of constraints.
The weighted precision is the average of the precision calculated for each class of conformance, i.e., if the rule should be fulfilled or not. 
The intuition of employing the \emph{Vanilla} model is to check if and how much of the original set of rules is being kept.
We can notice that for any template, this model performs quite randomly, since it keeps a median precision of 58\% across all the event logs, highlighting its stochastic nature. 
The \emph{Original} model, on the other hand, illustrates the capability of the model to learn the original set of rules, with a median precision of 99\% across all the event logs.
This approach can be considered a sanity check since we are replaying the original set of rules to the model and checking if it can preserve the original semantics of processes.

Although the \emph{Original} simulation is capable of fulfilling the imposed rules, we can notice that not all the set of inverted rules are being fully fulfilled by observing the \emph{Inverted Subset}.
This is particularly expected since the model is not trained to fulfill the inverted set of rules, but the original set of rules.
However, the achieved precision is remarkably high in general, with a median of 81\% across all the event logs.
This enforces the generalization capability presented by neural networks, which can learn the original set of rules along with the event attributes and apply them to a different set of rules, even if they differ from the original set of observed rules.
This is particularly important for the process simulation since it is not always possible to know the set of rules that will be imposed on the model in the future, and the model should be able to generalize the learned set of rules to a different set of rules.
However, for some templates and event logs, we can notice a higher difficulty in generalizing.
For instance, in the templates, $\mathcal{C}$ and $\mathcal{PR}$ on the \bpiseventeen~event log, the \emph{Original} approach has a slightly lower precision compared to the other logs, which is drastically reflected on the inverted set of rules.
This clearly identifies the difficulty of learning the rules and, hence, the inability to generalize to the unseen combination of event attributes and inverted rules. 

Furthermore, we can notice that some templates are harder than others. 
Table~\ref{tab:template-precision} shows the average precision for each class of conformant rules, i.e., to be or not to be fulfilled, for each template regarding the \emph{Inverted} simulation.
We can notice that the precision for rules that should not be fulfilled is higher and this behavior is due to a class imbalance issue, where the not fulfilled corresponds to the majority class in most cases.
More precisely, in our experimental setup, across all the event logs and templates, the number of traces corresponding to the set of rules to be fulfilled is $285\pm8$ total rules, while the set of rules to be not fulfilled corresponds to $1000\pm200$ total traces.
Finally, the difference in performances among the templates is intuitive due to the nature of each of those templates, where some of them are more complex than others due to their higher dimensionalities (i.e., more rules $\phi$), making it more difficult to be learned by the model.
We extend this discussion in the following.


\begin{wraptable}[9]{r}{6.3cm}
  \centering
  \scriptsize
  \caption{Precision for each class of rules.}
  \label{tab:template-precision}
  \begin{tabular}{@{}ccc@{}}
    \toprule
    \textbf{Template}           & \textbf{Fulfilled} & \textbf{Not fulfilled} \\ \midrule
    \textbf{Existence}          & 0.912 ± 0.2                  & 0.917 ± 0.18                 \\
    \textbf{Choice}             & 0.807 ± 0.28                 & 0.9048 ± 0.15                \\
    \textbf{Pos. Relations} & 0.6468 ± 0.29                & 0.9053 ± 0.12                \\ \bottomrule
    \end{tabular}%
\end{wraptable} 

\begin{figure}
  \centering
  \begin{subfigure}[b]{0.59\linewidth}
      \centering
      \includegraphics[width=\linewidth]{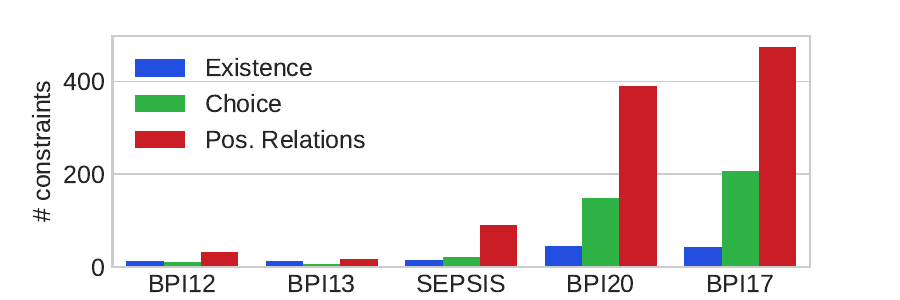}
      \caption{}
      \label{fig:n-constraints-log}
  \end{subfigure}
  \hfill
  \begin{subfigure}[b]{0.39\linewidth}
      \centering
      \includegraphics[width=\linewidth]{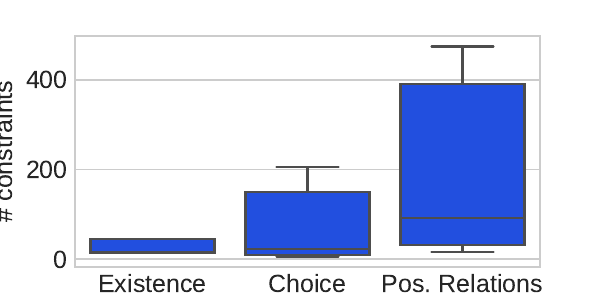}
      \caption{}
      \label{fig:n-constraints-template}
  \end{subfigure}
  \caption{Analysis of event log constraints from two perspectives: (a) count per log across templates, (b) distribution per template.}
  \label{fig:log-characteristics}
\end{figure}

\textbf{Log and DECLARE templates characteristics}. The difficulty of learning to satisfy imposed constraints can be also reflected in the number of constraints for each event log under a certain template.
We illustrate the number of constraints obtained by each template and for each event log in Figure~\ref{fig:log-characteristics}. 
More specifically, Figure~\ref{fig:n-constraints-log} shows the number of constraints for each event log under different templates, whereas Figure~\ref{fig:n-constraints-template} shows the distribution of the number of constraints for each template.
We can notice that the $\mathcal{PR}$ template is the one with the highest number of constraints, which results in a higher dimensional feature vector for the conditioned simulation.
Intuitively, the higher the dimensionality of the feature vector, the harder it is for the model to learn the set of rules, and the harder it is for the model to generalize the learned set of rules to a different set of rules.
Additionally, in Figure~\ref{fig:weighted-precision} the \bpiseventeen~had the poorest performance, which is reflected in the higher number of constraints for the $\mathcal{PR}$ template, as depicted in Figure~\ref{fig:n-constraints-log}. 
Furthermore, despite the similar characteristics with the \bpipermit, the \bpiseventeen~is even more complex due to its longer traces.
It is well known by the deep learning community that longer sequences are harder to learn and generalize, specifically for recurrent neural networks.

Besides the log characteristics, another factor that impacts the ability of our proposal to generalize to different sets of rules is the selected activities to simulate.
Recall that the activities illustrated in Table~\ref{tab:simulated-templates} were selected based on the rule frequencies, which can result in a bottleneck.
Due to the lack of space, we are not able to illustrate the process models of the event logs, but as a brief discussion, we noticed that for the \bpiseventeen~under the template $\mathcal{E}$, the simulated activity \emph{W Handle leads} has several self-loops, a characteristic not found in the other event logs under the same template.
Another characteristic observed from the process models is that the \emph{End Tripe} and \emph{Send Reminder} activities are far away from each other in the \bpipermit, which also indicates a more complex semantic to be captured.

\begin{figure}
  \centering
  \includegraphics[width=.9\textwidth]{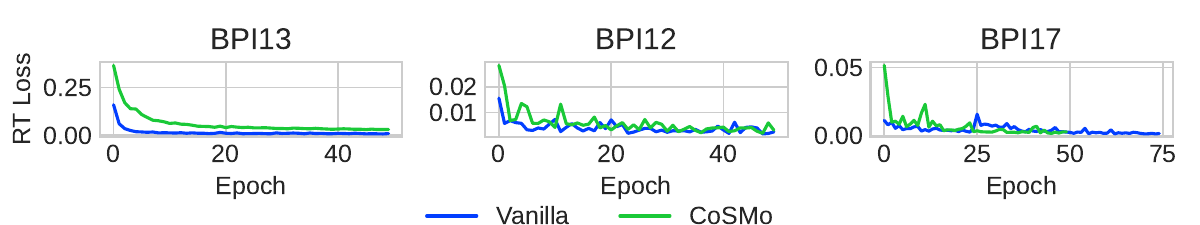}
  \caption{Data-flow of the proposed CoSMo framework.}
  \label{fig:remaining-time}
\end{figure}

\textbf{Data-flow}. As a final discussion, we illustrate the model capability of auto-regressively simulating the remaining time along with the activities under the imposed constraints versus the vanilla approach. 
The performance of the remaining time (RT) is illustrated as the loss function (MAE) in Figure~\ref{fig:remaining-time}.
Due to the limited space, we are illustrating only three event logs with varying characteristics: the \bpiproblems, the \emph{BPI12}, and the \bpiseventeen, respectively illustrating a small, medium, and large event log.
We can notice a slower convergence for CoSMo and this is due to the fact we are increasing the complexity of the model by including the constraints to be learned. 
However, around 20 epochs both the Vanilla and CoSMo can converge and achieve a similar performance.
We are not diving into this discussion but in the future, we plan to derive the timestamps from the predicted remaining time and employ relevant metrics to better assess the quality of these simulations.
Moreover, it is not novel that neural networks perform very well for this specific task, but we include this brief discussion to ensure that the remaining time prediction also performs well under the constrained scenario proposed in this work.

\color{black}

\subsection{Distinctive Advancements and Limitations}

Our research innovates in process simulation using deep learning by moving beyond the conventional stochastic generative models, focusing instead on an architecture that integrates extensive constraints and a-priori knowledge for a nuanced understanding and simulation of data patterns, diverging from mere sequence prediction.
Unlike conventional deep learning solutions~\cite{CamargoDR21}, which are predominantly stochastic generative models (as elaborated in section~\ref{sec:related}), our approach takes a step towards overcoming this limitation. 
Existing methods focus primarily on sequence prediction, which is often similar to, but not synonymous with, simulation. 
In contrast, a key innovation in CoSMo's inference stage is its ability to generalize and adaptively manipulate process changes through the constraint vector, represented in this study by \ltlf~rules. 
This feature represents a significant step toward addressing the stochastic nature inherent in deep learning models and incorporating critical information into the learned model to allow more sophisticated evaluations (e.g., what-if analysis).
Such capabilities are not only essential for PM practitioners and stakeholders but are also ideal for complex situations where a deep understanding and adherence to an extensive set of rules or a-priori knowledge is essential.

While our study has demonstrated promising outcomes, it is important to acknowledge certain limitations to be addressed in future research. 
First, the application of the DECLARE language modeling in our case study highlighted some bottlenecks, notably the correlation among some rules (e.g., existence and absence rules), which might introduce redundant information into the learning process. 
To enhance model efficiency and accuracy, future work will explore preprocessing steps to eliminate these correlated features, thereby streamlining the feature vector and potentially improving performance on complex templates and datasets.
Secondly, our focus has been predominantly on methodological validation, with less attention to the architecture's applicability in real-world settings. 
The process of rules and activity selections (see Table~\ref{tab:simulated-templates}), conducted arbitrarily in this study, would benefit from the involvement of domain experts for each event log in practical scenarios to ensure the relevance and impact of chosen rules on the simulations are accurately assessed.
Lastly, there is a need to broaden the scope of exploration to include diverse types of constraints and a-priori knowledge, moving beyond declarative rules to incorporate, for instance, user-based constraints (such as time-related constraints to change, for instance, the distribution of waiting time) and intra-case feature availability (e.g., resources). 

\section{Conclusion}\label{sec:conclusion}

In this work, we introduced a novel recurrent architecture tailored for conditioned process simulation, embodied in the CoSMo framework. 
Through evaluations of real-world event logs and DECLARE templates, CoSMo demonstrated its ability to adapt to various constraints and datasets, offering process simulations that align with specified conditions. 
Moreover, CoSMo mitigates the inherent stochasticity of deep learning models, enhancing its suitability for process simulation. 
Acknowledging the discussed limitations, our findings pave the way for future advancements that could significantly improve the application of deep learning-based process simulation models.
Future works include the exploration of additional constraints and a-priori knowledge, practical applications via what-if analysis, the involvement of domain experts for rule selection, and the inclusion of other event attributes to be simulated.

\begin{credits}
\subsubsection{\ackname} Supported by Università degli Studi di Milano (PON DM 1061/ 2021, action IV.5, FSE REACT-EU CUP: G45F21002100006), LMU Munich, and the Munich Center for Machine Learning.

\end{credits}

\bibliographystyle{splncs04}
\bibliography{bibliography}

\begin{thebibliography}{10}
\providecommand{\url}[1]{\texttt{#1}}
\providecommand{\urlprefix}{URL }
\providecommand{\doi}[1]{https://doi.org/#1}

\bibitem{Aalst15}
van~der Aalst, W.M.P.: Business process simulation survival guide. In: Handbook
  on Business Process Management, 2nd Ed, pp. 337--370. International Handbooks
  on Information Systems, Springer (2015)

\bibitem{Alman24declare4py}
Alman, A., Donadello, I., Maggi, F.M., Montali, M.: Declarative process mining
  for software processes: The rum toolkit and the declare4py python library.
  In: Product-Focused Software Process Improvement. pp. 13--19 (2024)

\bibitem{AlmanMMP22pdeclare}
Alman, A., Maggi, F.M., Montali, M., Pe{\~{n}}aloza, R.: Probabilistic
  declarative process mining. Inf. Syst.  \textbf{109},  102033 (2022)

\bibitem{Burattin15}
Burattin, A.: {PLG2:} multiperspective processes randomization and simulation
  for online and offline settings. CoRR  \textbf{abs/1506.08415} (2015)

\bibitem{BurattinRRT22}
Burattin, A., Re, B., Rossi, L., Tiezzi, F.: A purpose-guided log generation
  framework. In: {BPM}. vol. 13420, pp. 181--198. Springer (2022)

\bibitem{CamargoDG20}
Camargo, M., Dumas, M., Gonz{\'{a}}lez, O.: Automated discovery of business
  process simulation models from event logs. Dec. Sup. Sys  \textbf{134},
  113284 (2020)

\bibitem{CamargoDR19}
Camargo, M., Dumas, M., Rojas, O.G.: Learning accurate {LSTM} models of
  business processes. In: {BPM}. vol. 11675, pp. 286--302. Springer (2019)

\bibitem{CamargoDR19b}
Camargo, M., Dumas, M., Rojas, O.G.: Simod: {A} tool for automated discovery of
  business process simulation models. In: {BPM} Workshop. vol.~2420, pp.
  139--143 (2019)

\bibitem{CamargoDR21}
Camargo, M., Dumas, M., Rojas, O.G.: Discovering generative models from event
  logs: data-driven simulation vs deep learning. PeerJ CS  \textbf{7}, ~e577
  (2021)

\bibitem{CamargoDR22hybridsim}
Camargo, M., Dumas, M., Rojas, O.G.: Learning accurate business process
  simulation models from event logs via automated process discovery and deep
  learning. In: CAiSE. pp. 55--71. Springer (2022)

\bibitem{Donadello23oodeclare}
Donadello, I., {Di Francescomarino}, C., Maggi, F.M., Ricci, F., Shikhizada,
  A.: Outcome-oriented prescriptive process monitoring based on temporal logic
  patterns. {EAAI} journal  \textbf{126} (2023)

\bibitem{Dumas21}
Dumas, M.: Constructing digital twins for accurate and reliable what-if
  business process analysis. In: {BPM} Workshop. vol.~2938, pp. 23--27 (2021)

\bibitem{Estrada21ddslimitation}
Estrada{-}Torres, B., Camargo, M., Dumas, M., Garc{\'{\i}}a{-}Ba{\~{n}}uelos,
  L., Mahdy, I., Yerokhin, M.: Discovering business process simulation models
  in the presence of multitasking and availability constraints. Data Knowl.
  Eng.  \textbf{134},  101897 (2021)

\bibitem{Francescomarino17}
Francescomarino, C.D., Ghidini, C., Maggi, F.M., Petrucci, G., Yeshchenko, A.:
  An eye into the future: Leveraging a-priori knowledge in predictive business
  process monitoring. In: {BPM}. vol. 10445, pp. 252--268. Springer (2017)

\bibitem{GiacomoV13ltl}
Giacomo, G.D., Vardi, M.Y.: Linear temporal logic and linear dynamic logic on
  finite traces. In: {IJCAI}. pp. 854--860. {IJCAI/AAAI} (2013)

\bibitem{MartinDC16}
Martin, N., Depaire, B., Caris, A.: The use of process mining in business
  process simulation model construction - structuring the field. Bus. Inf.
  Syst. Eng.  \textbf{58}(1),  73--87 (2016)

\bibitem{MeneghelloFG23rims}
Meneghello, F., Francescomarino, C.D., Ghidini, C.: Runtime integration of
  machine learning and simulation for business processes. In: {ICPM}. pp. 9--16
  (2023)

\bibitem{Pasquadibisceglie19}
Pasquadibisceglie, V., Appice, A., Castellano, G., Malerba, D.: Using
  convolutional neural networks for predictive process analytics. In: {ICPM}.
  pp. 129--136. {IEEE} (2019)

\bibitem{PesicSA07declare}
Pesic, M., Schonenberg, H., van~der Aalst, W.M.P.: {DECLARE:} full support for
  loosely-structured processes. In: {EDOC}. pp. 287--300 (2007)

\bibitem{ManeiroVL21}
Rama-Maneiro, E., Vidal, J., Lama, M.: Deep learning for predictive business
  process monitoring: Review and benchmark. IEEE TSC pp.~1--1 (2021)

\bibitem{RosingWCM15bpmn}
von Rosing, M., White, S., Cummins, F., de~Man, H.: Business process model and
  notation - {BPMN}. In: The Complete Business Process Handbook, pp. 429--453.
  Morgan Kaufmann/Elsevier (2015)

\bibitem{RozinatMSA09}
Rozinat, A., Mans, R.S., Song, M., van~der Aalst, W.M.P.: Discovering
  simulation models. Inf. Syst.  \textbf{34}(3),  305--327 (2009)

\bibitem{SaniGZA20}
Sani, M.F., Gonzalez, J.J.G., van Zelst, S.J., van~der Aalst, W.M.P.:
  Conformance checking approximation using simulation. In: {ICPM}. pp.
  105--112. {IEEE} (2020)

\bibitem{TaxVRD17}
Tax, N., Verenich, I., Rosa, M.L., Dumas, M.: Predictive business process
  monitoring with {LSTM} neural networks. In: CAiSE. vol. 10253, pp. 477--492.
  Springer (2017)

\bibitem{WeytjensW21a}
Weytjens, H., Weerdt, J.D.: Creating unbiased public benchmark datasets with
  data leakage prevention for predictive process monitoring. In: {BPM}.
  vol.~436, pp. 18--29. Springer (2021)

\end{thebibliography}

\end{document}